\documentclass{article}

     \usepackage[final, nonatbib]{neurips_2019}

\usepackage[utf8]{inputenc} 
\usepackage[T1]{fontenc}    
\usepackage{hyperref}       
\usepackage{url}            
\usepackage{booktabs}       
\usepackage{amsfonts}       
\usepackage{nicefrac}       
\usepackage{microtype}      
\usepackage{graphicx}
\usepackage{xcolor}
\usepackage{subcaption}
\usepackage{multirow}
\usepackage{wrapfig}
\usepackage{adjustbox}
\usepackage[ruled,vlined,noresetcount]{algorithm2e}
\title{Accelerating Training in Pommerman with Imitation and Reinforcement Learning}

\author{Hardik Meisheri \\ TCS Research \\ Mumbai, India \\ \textit{hardik.meisheri@tcs.com} \And Omkar Shelke \\ TCS Research \\ Mumbai, India \\ \textit{shelke.omkar@tcs.com} \And Richa Verma \\ TCS Research \\ Delhi, India \\ \textit{richa.verma4@tcs.com} \And Harshad Khadilkar \\ TCS Research \\ Mumbai, India \\ \textit{harshad.khadilkar@tcs.com}
}

\begin{document}
	
	\maketitle
	
	\begin{abstract}
		The Pommerman simulation was recently developed to mimic the classic Japanese game Bomberman, and focuses on competitive gameplay in a multi-agent setting. We focus on the 2$\times$2 team version of Pommerman, developed for a competition at NeurIPS 2018\footnote{https://nips.cc/Conferences/2018/CompetitionTrack}. Our methodology involves training an agent initially through imitation learning on a noisy expert policy, followed by a proximal-policy optimization (PPO) reinforcement learning algorithm. The basic PPO approach is modified for stable transition from the imitation learning phase through reward shaping, action filters based on heuristics, and curriculum learning. The proposed methodology is able to beat heuristic and pure reinforcement learning baselines with a combined 100,000 training games, significantly faster than other non-tree-search methods in literature. We present results against multiple agents provided by the developers of the simulation, including some that we have enhanced. We include a sensitivity analysis over different parameters, and highlight undesirable effects of some strategies that initially appear promising. Since Pommerman is a complex multi-agent competitive environment, the strategies developed here provide insights into several real-world problems with characteristics such as partial observability, decentralized execution (without communication), and very sparse and delayed rewards. 
		
	\end{abstract}
	
	\section{Introduction}
	
	Reinforcement learning has achieved success in solving several complex problems, ranging from game playing \cite{mnih2015human,silver2016mastering} to robotics \cite{pmlr-v87-matas18a} and autonomous driving \cite{shalev2016safe}. Many algorithms originally developed for gameplay have been subsequently adapted for real-world applications, highlighting the importance of the former from both theoretical and practical perspectives. However, many of the current algorithms in RL have been designed for single-agent domains, where the environment is either stationary \cite{mirowski2016learning}, or else is subject to a fixed set of rules or policies \cite{mnih2015human}. In addition, RL algorithms are prone to sample inefficiency, due to which it takes vast amount of training to reach to desirabele performance~\cite{yu2018towards}. Relatively few studies \cite{silver2016mastering} have considered situations with human or AI-driven opponents. Building RL algorithms for \textit{mixed} cooperative and competitive environments with complex dynamics is difficult, because of the challenge of separating the true reward signal from noise. At the same time, many real-world applications such as multi-robot exploration \cite{matignon2012coordinated} and auctions \cite{nanduri2007reinforcement} make this problem interesting from a practical standpoint in addition to its theoretical depth.
	
	The key challenges in multi-agent scenarios are as follows. First, non-stationarity of the environment from the perspective of any single agent means that not all rewards are explainable by changes in the agent’s own policy \cite{oliehoek2016concise}. This also leads to another problem of credit assignment among the agents when there are sparse and common rewards \cite{minsky1961steps}. Second, environments such as Pommerman can impose restrictions on communication\footnote{There was no inter-agent communication in the NeurIPS 2018 competition, while two bits of information can be exchanged in each time step for the NeurIPS 2019 version.}, which disqualifies multi-agent RL approaches with centralised critics. Restricted communication is not peculiar to Pommerman, but can be found in several practical situations such as drone swarms as well. Third, constraints such as partial observability and sparse rewards further increase the complexity of the problem, leading to the possibility of policy degeneration.
	
	Two approaches from prior literature that address these issues are to either roll out the environment through tree search \cite{kartal2018using,kartal2019safer} or to undertake extensive training \cite{peng2018continual,malysheva2018deep}, both of which require significant computational resources. In this paper, we aim to strike a balance between the purity of from-scratch RL policy search, with the limitations of imitating a noisy expert policy. We do so by initially imitating the noisy expert policy (a simple heuristic provided by the game developers) in order to learn the basic functionality of Pommerman \cite{resnick2018pommerman}, and follow this by training using a stochastic on-policy algorithm. The key contributions of this paper are, (i) a stable learning paradigm for imitation followed by RL-driven improvements without allowing policy forgetting, (ii) a significant reduction in training duration compared to prior literature, and (iii) extensive evaluation of the proposed method in terms of behaviour as well as performance against baseline agents.
	
	\begin{wrapfigure}{o}{0.5\textwidth}
		\centering
		\includegraphics[width = 0.85\linewidth, keepaspectratio]{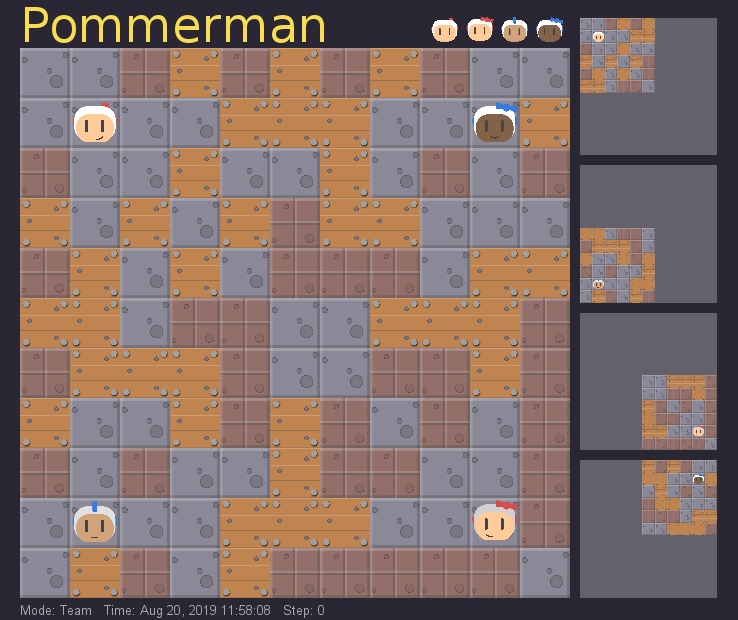}
		\caption{Sample initial board layout. Visibility for each agent is shown in the panel on the right.}
		\label{fig:sample_board}
	\end{wrapfigure}
	
	\textbf{About Pommerman: }
	The basic Pommerman environment contains three variants: FFA (free for all, a fully observable mode with a single player against 3 opponents), Team (the partially observable 2$\times$2 mode that we consider in this paper), and TeamRadio (team variant with communication). The Team environment contains an $11\times11$ board with agents spawning at each corner, with teammates starting in opposite corners as shown in Figure~\ref{fig:sample_board}. At any given time, Agent can only see 5 cells from its position in any direction. The objective of the game is to survive and to kill the opponents by placing bombs. Bombs explode 10 time ticks after placement. Flames from the bomb last for 2 time ticks. Initially, the bomb blast range is 3 in horizontal as well as vertical direction. There are 6 discrete actions, 4 for cardinal movement and 1 each for placing a bomb and doing nothing. In addition, there are power-ups which can increase the blast radius of the bomb, increase ammo capacity to place more than one bomb simultaneously, and the capability to kick bombs away. There are two types of walls, wooden and stone. Wooden walls can be destroyed by the bombs and might reveal power-ups, whereas stone walls are unaffected. Agents can only move where there are passages. Each game starts with a random generation of stone walls and wooden walls, which are symmetric along the diagonal. If neither of the team is able to win after 800 time ticks, the game is said to be tied. Each agent has partial visibility of 5 cells in each direction.
	
	\section{Related Work}
	
	
	The unique challenges in Pommerman have attracted many researchers to this environment. Their approaches can be broadly categorized into model-free RL \cite{hernandez2019agent, resnick2018backplay, peng2018continual, malysheva2018deep, gao2019skynet} and tree-search-based-RL \cite{perez2019analysis, zhou2018hybrid, kartal2018using, osogami2019real, kartal2019safer}. In addition, \cite{shahmulti} is an excellent review of Pommerman, its practical implications, and its limitations. A comparison of search techniques including MCTS, breadth-first, and flat Monte Carlo \cite{zhou2018hybrid} shows that in the fully observable FFA mode, MCTS is able to beat simpler and hand-crafted solutions. An extension of this study \cite{perez2019analysis} called Rolling Horizon Evolutionary Algorithm (RHEA) concludes that the more offensive strategies (like RHEA with a high rate of bomb placing) are normally also riskier, due to inadvertent suicides\footnote{This is also visible in one of our agents when trained in a raw manner without curriculum learning}. One way around this is to perform tree search using pessimistic scenarios \cite{osogami2019real}, and to choose actions that minimise the risk. Since the worst scenario can be deterministic, it can be rolled out efficiently. However, unrealistic or illegal scenarios can be generated and these have a detrimental effect on learning.
	
	Studies that propose prediction of the movements of the other agents in addition to learning self policy \cite{hernandez2019agent} are based on the hypothesis that this would improve coordination in multi-agent scenarios. Continual learning \cite{peng2018continual} was used to train a population of advantage-actor-critic (A2C) agents in Pommerman, beating all other learning agents in the 2018 Competition. A Deep Neural Network (DNN) is updated using A2C in a process that allows the agent to progressively learn new skills, such as picking items and hiding from bomb explosions. Another Deep Learning approach is proposed by \cite{malysheva2018deep}, which uses Relevance Graphs obtained by a self-attention mechanism. This agent, enhanced with a message generation system, analyses the relevance of other agents and items observed in the environment. Backplay \cite{resnick2018backplay} speeds up training by backtracking from the terminal states to the initial states of episodes, improving sample-efficiency. \textit{Skynet} \cite{gao2019skynet} trains deep neural networks using Proximal Policy Optimization (PPO). They have also implemented reward shaping and trained using curriculum learning paradigm. This the closest to our work, however, they do not employ imitation learning and train the network using PPO from scratch, which requires tremendous amount of training and compute. In \cite{kartal2018using}, also later expanded in \cite{kartal2019safer}, the authors train a DNN using Asynchronous Advantage Actor-Critic (A3C) enhanced with temporal distance to goal states. They also integrate MCTS as a demonstrator for A3C, which helps reduce agent suicides during training via imitation.
	
	\section{Proposed approach} \label{sec:method}
	
	The problem can be modeled as a markov decision process, ($\mathcal{S}, \mathcal{A}, \mathcal{T}, \mathcal{R}, \gamma$), where $\mathcal{S}$ represents the partially observable state, $\mathcal{A}$ denotes the six actions, $\mathcal{T}$ represents transition probabilities and $\mathcal{R} $ denotes reward. Our focus in this paper is on a model-free approach and hence the transition probabilities are not modeled. A potential way of reducing the computational effort for training is to use off-policy sample-efficient algorithms such as DQN \cite{mnih2015human}. However, the partial observability, sparse reward structure, and long episode length (up to 800 steps) make it difficult to use experience replay for stabilising deep Q-learning. At the other end of the spectrum, simpler on-policy methods such as policy gradient are susceptible to high variance. Therefore we turn to methods based on the actor-critic architecture. Trust Region Policy Optimization (TRPO) \cite{schulman2015trust} maximizes an objective function similar to vanilla policy gradient method, subject to a constraint on the size of policy update. However, TRPO needs a second order derivative to compute gradients and hence, is very computationally expensive. Proximal Policy Optimization (PPO) \cite{schulman2017proximal} achieves similar performance while relying on first order derivative and hence is more efficient. We use PPO in our approach.
	
	\subsection{State Space and Network Architecture}
	
\begin{figure}[b]
		\vskip-15pt
		\centering
		\includegraphics[width=0.55\linewidth]{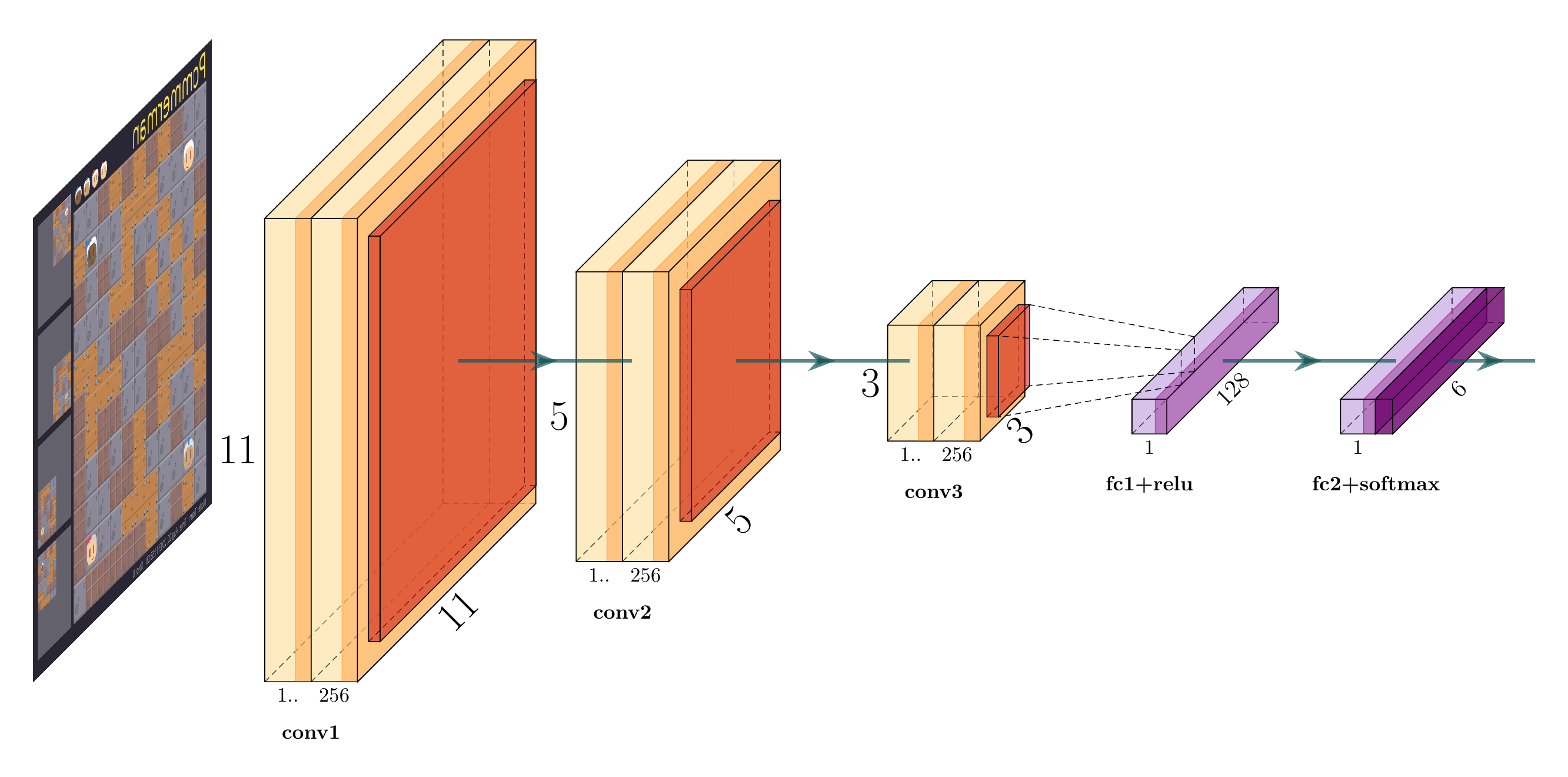}
		\vskip-10pt
		\caption{CNN Architecture for Policy}
		\label{fig:network_architecture}
	\end{figure}

	Pommerman environment provides observation in a dictionary in which, along with a board matrix of dimensions $11\times11$, we get other information such as the agents' bomb kicking capability, ammo, blast strength, IDs of two enemies and of the teammate at each time step. For our approach, we represent every feature as a separate $11\times11$ matrix which can be easily fed to a CNN. Categorical features such as items on the board are represented using a one-hot encoded matrix, whereas scalar features are populated as a full matrix. Apart from the raw information available in the input dictionary, we create one additional $11\times11$ input matrix representing the scalar \textit{desirability} of each observable tile on the board (for example, an open passage tile is more desirable than a bomb). This matrix is intended to encourage the agent to move towards desirable and safe positions on the board. In total, we get 19 channels in the input, details of which are given in Table~\ref{tab:features}. Our network comprises of three convolution layers, each with max pooling and dropouts followed by two fully connected layers. The output consists of six units with softmax activation, one for each action (Fig.~\ref{fig:network_architecture}). 
	
	\begin{table}[]
		\begin{tabular}{|l|l|}
			\hline
			
			\multirow{2}{*}{\parbox{2cm}{Board representation}} & \multirow{2}{*}{\parbox{11cm}{One channel each for one hot encoding of passage, rigid wall, wooden wall, bomb, flames, fog, extra bomb powerup, increase range powerup, kick powerup.}} \\
			\\
			\hline
			\multirow{2}{*}{\parbox{2cm}{Position encoding}} & \multirow{2}{*}{\parbox{11cm}{One channel each for the agent's own position, teammate's position and those of its enemies.}} \\ \\ \hline
			
			\multirow{2}{*}{\parbox{2cm}{Powerup representation}} & \multirow{2}{*}{\parbox{11cm}{A channel to broadcast the values of ammo, blast strength and binary kick capability.}} \\ \\ \hline
			
			\multirow{2}{*}{\parbox{2cm}{Bomb life and strength}} & \multirow{2}{*}{\parbox{11cm}{A channel to denote the blast strength and leftover lives of bombs placed on the board.}} \\ \\ \hline
			\multirow{3}{*}{\parbox{2cm}{Safe/desired cells}} & \multirow{2}{*}{\parbox{11cm}{The values of such cells are encoded as follows: Powerups=0, wooden wall=1, passage =2, Fog = 3, Enemies = 4, Rigid walls = 5, Teammate = 6, Bombs = 7, Flames = 8.}} \\ \\ \\ \hline
		\end{tabular}
		\caption{State space representation}
		\label{tab:features}
	\end{table}
	
	\subsection{Training Setup}
	
	We train first using imitation learning, followed by reinforcement learning, as mentioned in the introduction. The details are provided below, in addition to other modifications to the reward function and action selection.
	
	\textbf{Curriculum}: 
	The total training effort was equivalent to 150,000 games. Of these, the first 50,000 games were played purely using SimpleAgent (a default heuristic provided with the environment, described in detail later in Table~\ref{tab:agent_names}). State and action samples were saved for all four agents participating in each game. The network from Fig.~\ref{fig:network_architecture} is trained in supervised fashion with cross-entropy loss with states and actions as data instances and labels respectively.
	
	The imitation learning model acts as a policy network during the next training phase which uses PPO. A replica of the same network is created for value function estimation, with the output layer of size 1 instead of 6. We refrain from using the same CNN layers to approximate the value function as this creates aberration during the initial phase of learning and often leads to policy forgetting and degradation. We also avoid using any regularization technique such as dropout while training using PPO, as this leads to significant increase in KL divergence between trained policies. 
	
	The total effort with PPO is 100,000 games, played against agents of increasing sophistication (explained in Sec \ref{subsec:experiments}). We observed that training directly with the SimpleAgent or any other fully functional agent leads to forgetfulness of basic skills such as blasting wooden walls, picking powerups etc. In addtion, most drastic effect that leads to degradation of policies is learning to place bombs. This has been also observed in other studies \cite{kartal2019safer, resnick2018pommerman}. Training against agents with increasing difficulty helps retain the skills acquired in the imitation phase. 
	
	\textbf{Reward Shaping}: 
	The credit assignment problem can be broken down into two aspects. Assigning credit between the agents in a team, and for a single agent, distribution of rewards for different actions. The latter can be solved using generalized advantage estimates with a normalizing factor. Although the method is noisy, we observed stabilisation over the course of training.
	
	At the end of episode, we get only single reward for the team and it may not be clear how to assign credit to individual agents. For example, consider an episode where an agent eliminates an opponent but then commits suicide, and its teammate eliminates the remaining opponent. Under this scenario, both team members get a positive reward from the environment, but this could reinforce the suicidal behaviour of the first agent. Similarly, one agent could eliminate both opponents whereas its teammate just camps; both agents would get positive rewards, reinforcing a lazy agent~\cite{panait2005cooperative}. To solve the credit assignment problem within the team, we force the teammate to commit suicide at the start of each game (by placing a bomb and staying put until it explodes). Essentially, the game becomes 1-versus-2 and assigning the credit becomes easier. We provide a reshaped terminal reward signal as follows (at the risk of unintended changes in the policies~\cite{ng1999policy}). Note that we do not infer whether enemies died due to the PPO agent killing them, or through inadvertent suicides.
	
	\begin{itemize}
		\item Reward is -1 if the game ends with both enemies alive (no success, whether tied or lost)
		\item Reward is 0.5, if at the end of episode there is only enemy agent alive (could be a loss or a tie, but at least one enemy was killed)
		\item Reward is 1 if the agent wins (both enemies dead)
	\end{itemize}
	
	\textbf{Post-processing of selected actions:}
	We veto the actions chosen by PPO in two cases, in order to improve the training efficiency. We call these post-processing rules as \textit{jitter correction} and \textit{action filter}, and their motivation and definition is given below.
	
	\textit{Jitter Correction}: A peculiarity of the policies trained through imitation on SimpleAgent is a tendency to alternate between the same two actions in successive time steps (for example, right and left). This jittery behaviour is also observed in SimpleAgent itself, and the imitation learnt policy attaches very high confidence (nearly 1) to these actions. Therefore, the jitter is also inherited by PPO during the initial RL phase even though PPO is a stochastic algorithm. The behaviour is particularly noticeable when no enemies are visible to the agent, leading to there being no obvious objectives to achieve. A possible solution would be to use momentum-based approaches such as n-step predictions. However, they are not tested with partial observations and dynamic state spaces~\cite{lakshminarayanan2017dynamic, sharma2017learning}. Instead, we include a mechanism of jitter correction to break the agent out of its loop (Algorithm~\ref{algo:jitter_correction}).
	
	\begin{algorithm}[H]
		\SetAlgoLined
		xposition, yposition = empty list, empty list\\
		\While{not done}{
			append xpos, ypos to xposition, yposition\\
			static\_cond1 =  true if len(xposition[-15:]) ==1 else false\\
			static\_cond2 =  true if len(yposition[-15:]) ==1 else false\\
			
			x\_cond\_odd = true if len(set(xposition[-10::2])) == 1 else false \\
			x\_cond\_even = true if len(set(xposition[-11::2])) == 1 else false\\
			
			x\_cond\_uneq = false if  len(set(xposition[-11::2]) - (set(xposition[-10::2]))) == 0 else true\\
			
			x\_cond\_long = true if len(set(xposition[-35:])) == 2 else false\\
			x\_y\_cond\_long = true if len(set(xposition[-35:])) == 1 else false\\
			
			similar for y coordinate\\
			
			\lIf{static\_cond1 and static\_cond2}{
				Take next 3 steps from expert policy
			}\lElseIf{(x\_cond\_odd and x\_cond\_even and x\_cond\_uneq) or (x\_cond\_long and x\_y\_cond\_long)}{
			Take next 2 steps from expert policy
		}\lElseIf{ (y\_cond\_odd and y\_cond\_even and y\_cond\_uneq) or (y\_cond\_long and y\_x\_cond\_long)}{
		Take next 2 steps from expert policy
	}
}
\caption{Jitter Correction}
\label{algo:jitter_correction}
\end{algorithm}

\textit{Action Filter:} 
As observed in \cite{kartal2019safer}, there is a significant probability of an agent committing suicides at some point in the game, even with training. Avoiding this is particularly difficult because there are situations where the only way to avoid dying is to follow a long sequence of steps. We use a post-processing filter on the PPO actions in order to train efficiently (since the agent's death terminates the episode otherwise). This allows the agent to focus on higher level strategies. The PPO action is rejected if it is determined that the action would lead to death (for example, stepping into a bomb's path in the last time tick). Instead, any action apart from the PPO action and the bomb is chosen uniformly randomly. Given that the new action itself may be suicidal, the filter is applied until a safe action is found. A subtle difference between this approach and that of specifying the `correct' action, is that random choice allows for greater policy exploration. The Action Filter is implemented by rules shown in algorithm~\ref{algo:actionfilter}.

\begin{algorithm}[H]
	\SetAlgoLined
	
	act = agent.act(obs)\\
	next state = get\_next\_state(obs, action)\\
	
	\While{next\_state in flames or blast radius with bomb life remaining as 2}{Restrict that action, take any random action from \{right, left, top, bottom\} - \{act\}}
	
	\caption{ActionFilter}
	\label{algo:actionfilter}
\end{algorithm}

\subsection{Experimentation} \label{subsec:experiments}

As outlined earlier, we begin the reinforcement learning portion of training with a policy network trained using imitation learning. However, the value network required for PPO does not reuse these weights. Instead, we freeze the policy network and train only the value network for 10,000 games against SimpleAgent (default heuristic provided by the developers). Following this, we train our model against three types of opponent teams with increasing sophistication. They are explained in Table \ref{tab:agent_names}. We start with 10,000 games against StaticAgent, which makes no moves whatsoever. This portion of training is used to learn how to approach and kill opponents by placing bombs near them. Next we train for 20,000 games against SimpleAgent, but without allowing it to place bombs. This helps the PPO agent learn how to follow and trap opponents, but restricting their bomb capability allows it to learn this skill quickly (by prolonging the games). Finally, we train for 60,000 games against the default SimpleAgent. The total training after imitation thus lasts 100,000 games. We have provided more detail about the rationale behind the curriculum in Sec. \ref{sec:results} and Fig. \ref{fig:behaviour}.

As the probabilities of the actions drawn from deterministic policies trained using imitation are very skewed, the entropy coefficient in the PPO surrogate objective has been kept to zero. Keeping it to the default value as mentioned in the original paper, leads to catastrophic forgetting and degradation of the learned skills.  The PPO algorithm, like TRPO works on incremental updates in the policies and theoretical improves with respect to its previous policy. This provides a challenge while training, as Jitter Removal and Action Filter deviate from the pure PPO policy, and the resulting KL divergence between policies can be high. Keeping the higher threshold for KL divergence would also lead to degradation in policies, although that is also a function of batch size. Instead, we reduced the policy deviations with a probabilistic intervention: for each batch, only 10\% trajectories had Jitter Correction active and 30\% had Action Filter active. This provided stable learning and consistency in the observed KL divergence.  We use $128$ batch size and clip ratio of $0.01$ during training.

We train two separate agents starting from the same imitation-learned policy, \\
1. PPO with curriculum, reward shaping, jitter correction and Action Filter termed as PPOAgent\\
2. Vanilla PPO without any intervention termed as PPOAgent\_Cautious (for reasons explained later)\\
In the next section, we test our learned agents against various agents discussed above, to gauge the improvements over the initial imitation learning and the default heuristics.

\begin{table}[]
	\begin{tabular}{|l|l|}
		\hline
		
		\multirow{2}{*}{*\_jitter} & \multirow{2}{*}{\parbox{10cm}{Removing jitter, where agent is either stuck on a single cell or is alternating between two cells.}} \\
		\\
		\hline
		\multirow{2}{*}{*\_action} & \multirow{2}{*}{\parbox{10cm}{Preventing suicidal actions, for example, whether the next action leads into a bomb path.}} \\ \\ \hline
		
		StaticAgent & Agent which does not move from initial position \\ \hline
		SimpleAgent & Heuristic agent provided by the competition organizers \\ \hline
		
		\multirow{2}{*}{Imitation} & \multirow{2}{*}{\parbox{10cm}{Agent learned from the observations collected from SimpleAgent in a supervised setting}} \\ \\ \hline
		\multirow{2}{*}{PPO} & \multirow{2}{*}{\parbox{10cm}{Agent trained using Curriculum learning with PPO , with warmup weights from imitation.}} \\ \\ \hline
		PPOAgent\_Cautious & Agent trained with PPO with initial weights from imitation \\ \hline
	\end{tabular}
	\vspace{0.1cm}
	\caption{Nomenclature of agents}
	\label{tab:agent_names}
	\vskip-20pt
\end{table}











\section{Results and Discussion} \label{sec:results}

\begin{figure}
	\centering
	\includegraphics[width=0.88\linewidth]{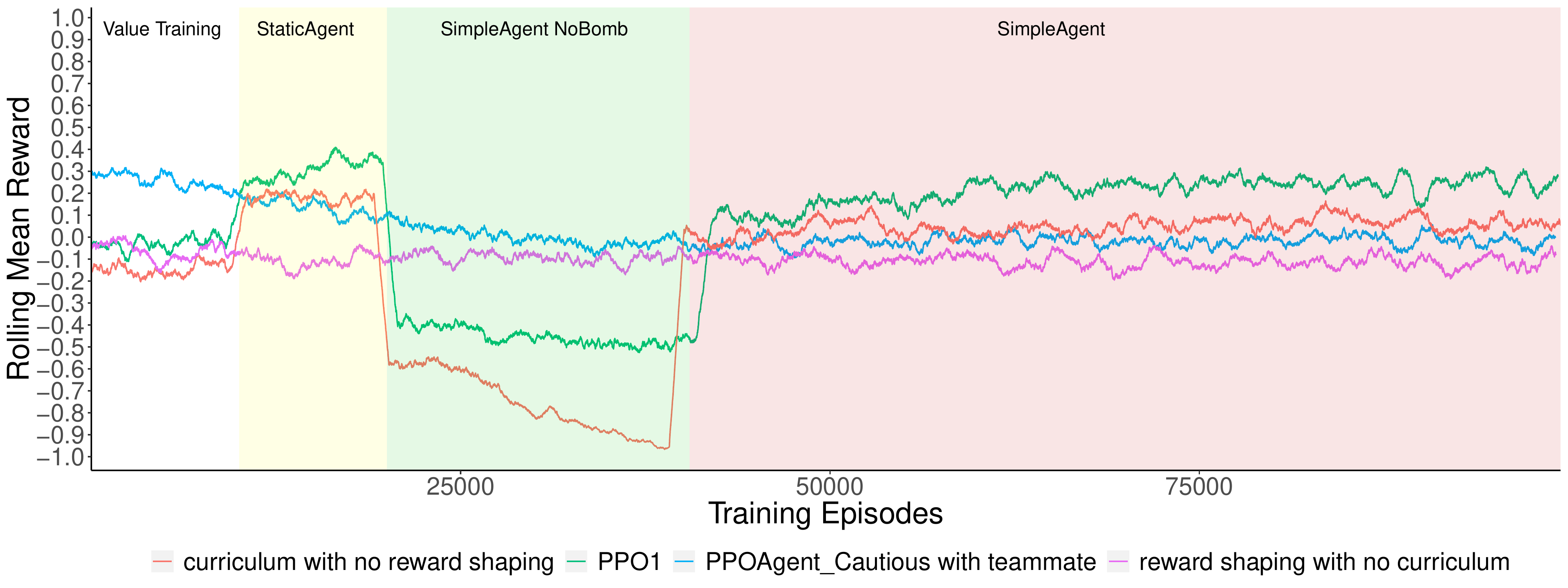}
	\caption{Reward during different training phases}
	\label{fig:training}
\end{figure}

We trained two agents for 100,000 games each: one (PPOAgent) with a curriculum of opponents, reward shaping, and post-processing of actions, and the other (PPOAgent\_Cautious) without any interventions. Fig.~\ref{fig:training} plots the evolution of training rewards for PPOAgent (green) and PPOAgent\_Cautious (blue). Two more agents are also shown: one that plays against the same curriculum of opponents without reward shaping (orange), and another that uses reward shaping but no curriculum of opponents (pink). Note that (i) the y-axis is a 1000-episode rolling mean of the rewards, and (ii) PPOAgent\_Cautious trains with its teammate (a SimpleAgent) active, so its initial reward is higher. Since the policy is invariant during the value training phase for PPOAgent, we know that this is the average reward for the imitation-learned policy against SimpleAgent. It is clear that the reward towards the end of training is higher than that for the imitation-learned policy, even though the plot only shows results with 10\% jitter correction and 30\% action filtering. The reward for PPOAgent\_Cautious reduces before stabilizing, probably due to credit assignment issues (its teammate is also active). Learning is slow even when the teammate is terminated (as in the pink curve), while both curriculum-based agents (blue and orange) show significantly faster progress. The agent with no reward shaping sees a -1 reward for both ties and losses, which makes it difficult to learn (see deterioration against SimpleAgent\_NoBomb). 
A visual rendering of both agents during gameplay shows that PPOAgent exhibits less jitter compared to its initial imitation-learned policy. Where it does enter a repetitive loop, it tends to do so in a finite area rather than just two neighbouring cells. This increases the probability of observing the enemy by accident, which breaks the loop. PPOAgent\_Cautious learns to avoid placing the bomb at all (hence the nomenclature), even for breaking wooden walls. This restricts its movement to the initial quadrant. Most wins for PPOAgent\_Cautious either due to an opponent committing suicide, or its teammate killing the opponents. Although it has a lower chance of accidentally dying than SimpleAgent, the learned policy returns very few wins.

\begin{figure}
	\centering
	\includegraphics[width=0.88\linewidth]{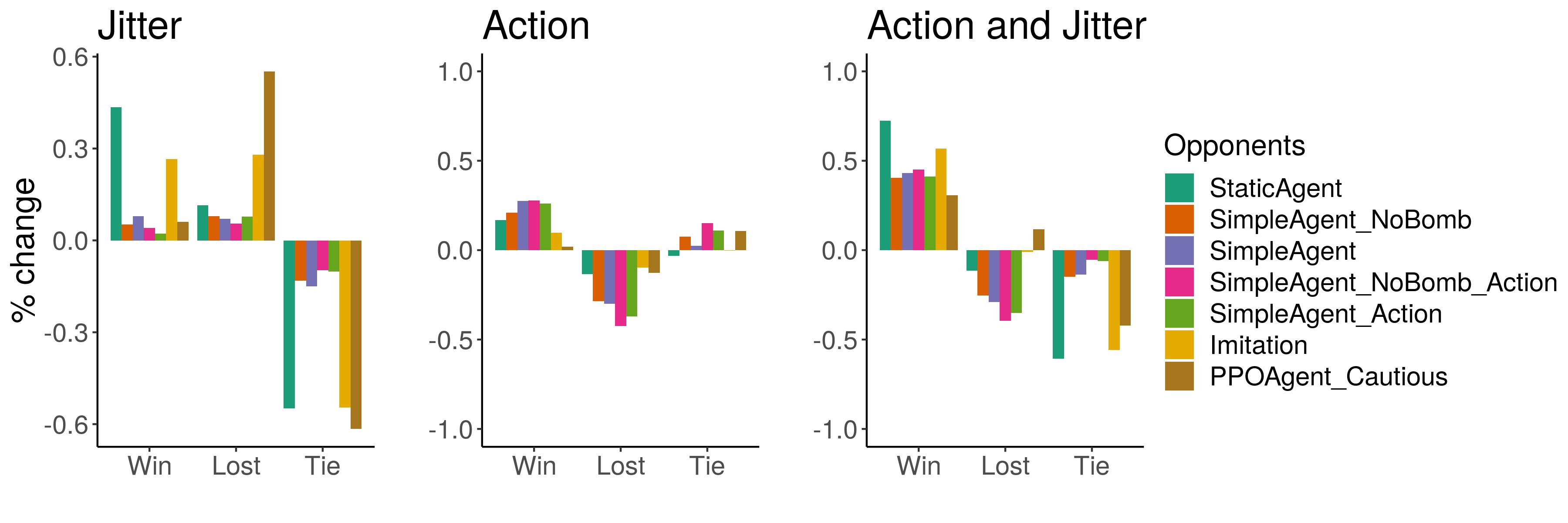}
	\caption{Effect of including Jitter and actionfilter}
	\label{fig:effect_jit_act}
\end{figure}

\begin{figure}
	\centering
	\begin{subfigure}{.5\textwidth}
		\centering
		\includegraphics[width = 0.8\linewidth, keepaspectratio]{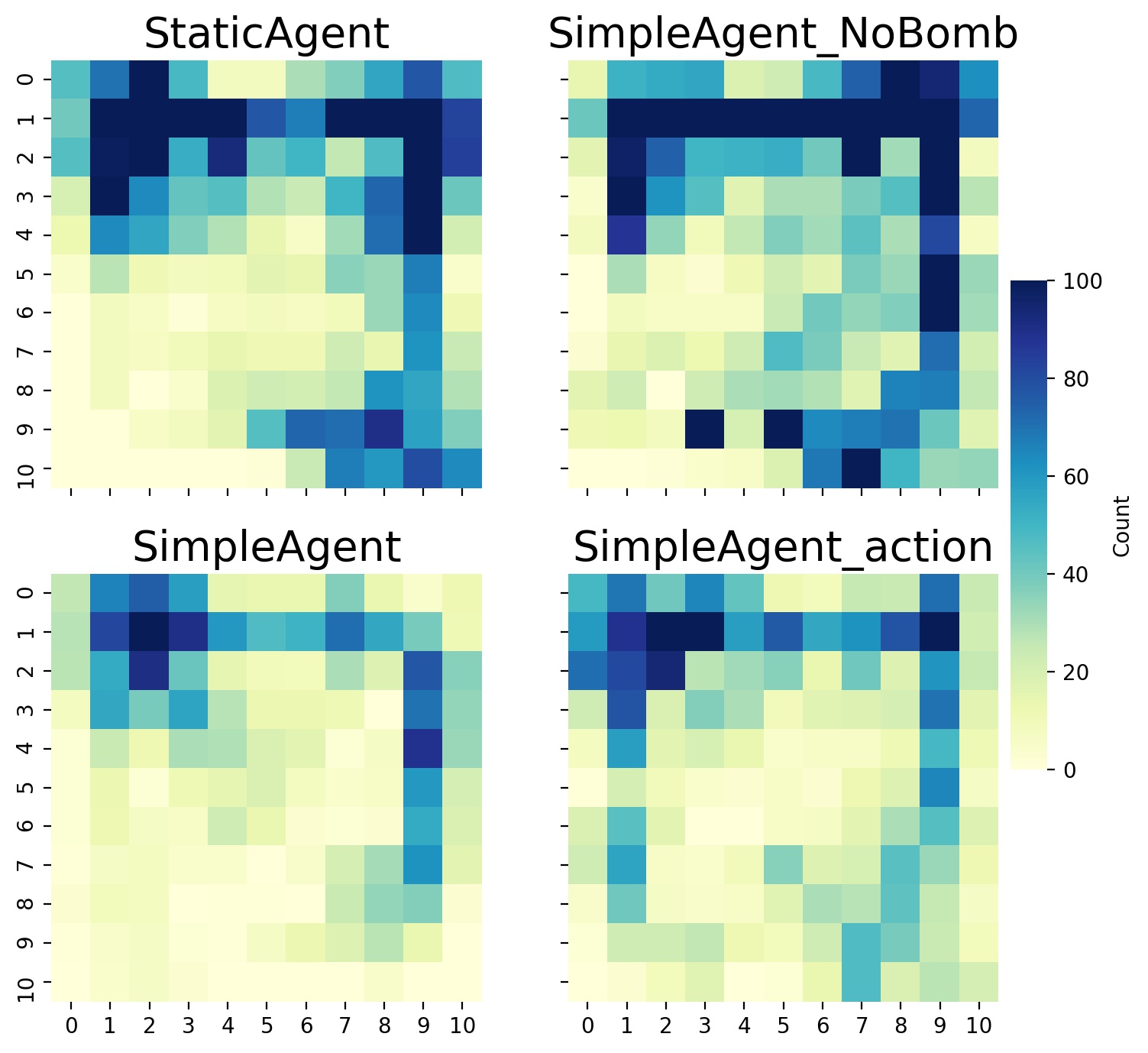}
		\caption{Exploration heatmap vs different agents}
		\label{fig:exploration}
	\end{subfigure}%
	\begin{subfigure}{.5\textwidth}
		\centering
		\includegraphics[width = 0.8\linewidth, keepaspectratio]{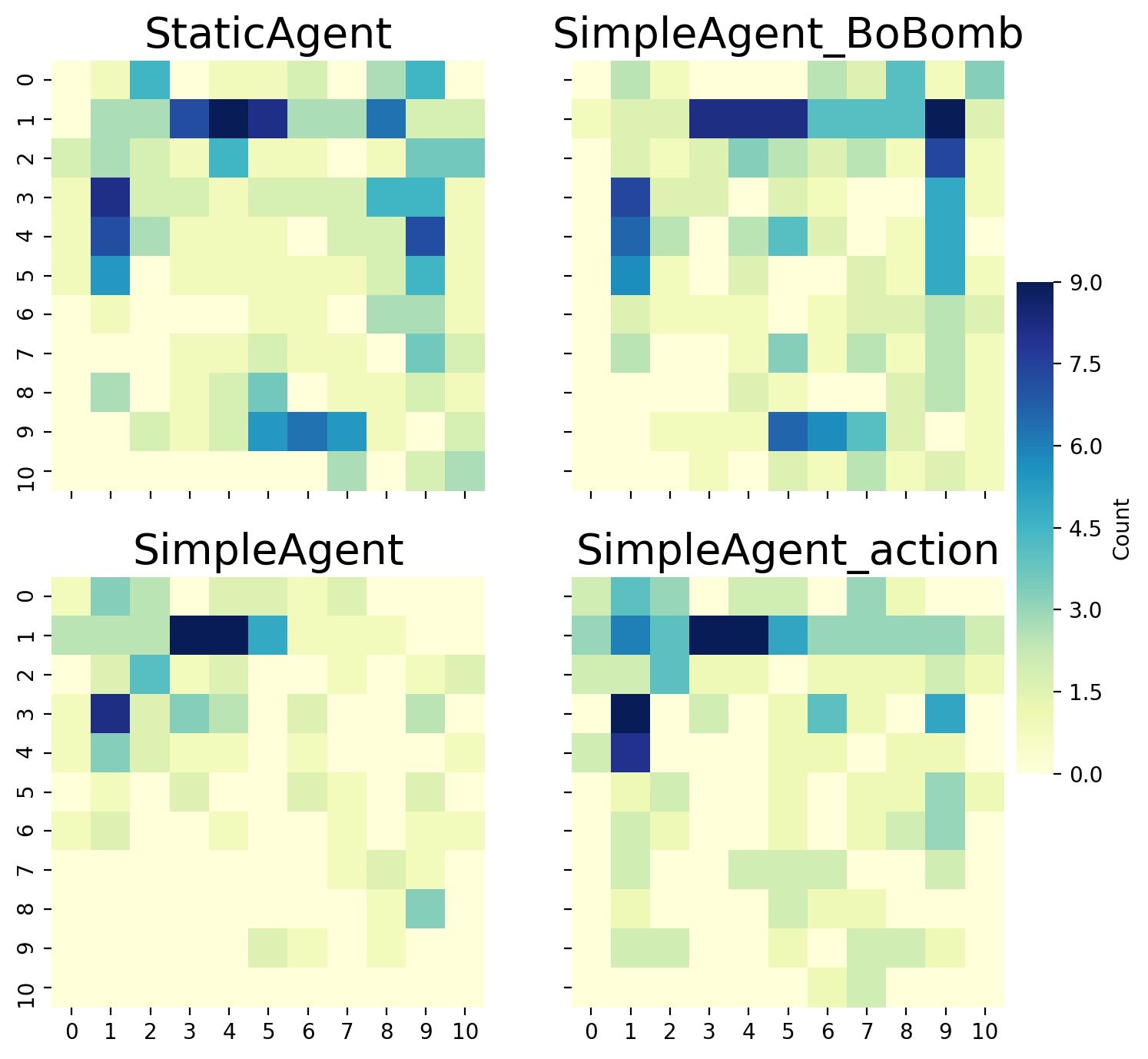}
		\caption{Bomb placement vs different agents}
		\label{fig:bomb_placement}
		
	\end{subfigure}
	\caption{Comparison of characteristics against different opponent strategies.}
	\label{fig:behaviour}
	\vskip-15pt
\end{figure}

Table \ref{tab:results_imitation} shows the performance of the agent trained using imitation learning on 50,000 games of SimpleAgent (initial policy used for PPO). The vanilla version uses the policy directly. Since we know that the policy is prone to jitter and to inadvertent suicides, we also test the policy augmented by one or both post-processing rules. These results act as a baseline for comparing the PPO results, which are given in Table~\ref{tab:results_ppo} (including those between PPO and imitation). The PPO results also include performance against Skynet\footnote{\url{https://github.com/MultiAgentLearning/playground/tree/NeurIPS-2018-Docker-Agents}}, which was the second best performing agent in the learning category in the 2018 NeurIPS competition. There are significant improvements over imitation, especially in the ratio of wins to losses. 
Furthermore, the PPO agent appears to win or tie 7 out of 8 games against Skynet.
\begin{table}[]
	\centering
	\begin{adjustbox}{max width=\textwidth}
		\begin{tabular}{@{}|l|l|l|l|l|l|l|l|l|l|l|l|l|@{}}
			\toprule
			\multirow{2}{*}{Opponents} & \multicolumn{3}{l|}{Imitation\_Vanilla} & \multicolumn{3}{l|}{Imitation\_jitter} & \multicolumn{3}{l|}{Imitation\_action} & \multicolumn{3}{l|}{Imitation\_jitter\_action} \\ \cmidrule(l){2-13} 
			& Win & Lost & Tie & Win & Lost & Tie & Win & Lost & Tie & Win & Lost & Tie \\ \midrule
			StaticAgent & 0.111 & 0.156 & 0.733 & 0.458 & 0.34 & 0.201 & 0.271 & 0.009 & 0.72 & 0.824 & 0.032 & 0.144 \\ \midrule
			SimpleAgent\_NoBomb & 0.355 & 0.416 & 0.299 & 0.379 & 0.507 & 0.114 & 0.615 & 0.109 & 0.275 & 0.753 & 0.115 & 0.131 \\ \midrule
			SimpleAgent & 0.331 & 0.418 & 0.251 & 0.361 & 0.498 & 0.141 & 0.603 & 0.099 & 0.297 & 0.756 & 0.126 & 0.118 \\ \midrule
			SimpleAgent\_NoBomb\_action & 0.218 & 0.607 & 0.175 & 0.242 & 0.663 & 0.095 & 0.493 & 0.176 & 0.331 & 0.63 & 0.227 & 0.143 \\ \midrule
			SimpleAgent\_action & 0.2 & 0.618 & 0.181 & 0.243 & 0.665 & 0.091 & 0.503 & 0.167 & 0.33 & 0.64 & 0.206 & 0.154 \\ \midrule
			PPO\_agent\_Cautious & 0.011 & 0.149 & 0.84 & 0.048 & 0.709 & 0.242 & 0.024 & 0.019 & 0.957 & 0.268 & 0.226 & 0.506 \\ \bottomrule
		\end{tabular}
	\end{adjustbox}
	\caption{Results in 1000 games for Imitation team (some games discarded due to fault after 12 steps).}
	\label{tab:results_imitation}
\end{table}
\begin{table}[]
	\begin{adjustbox}{max width=\textwidth}
		\begin{tabular}{@{}|l|l|l|l|l|l|l|l|l|l|l|l|l|@{}}
			\toprule
			\multirow{2}{*}{Opponents} & \multicolumn{3}{l|}{PPO\_Vanilla} & \multicolumn{3}{l|}{PPO\_jitter} & \multicolumn{3}{l|}{PPO\_action} & \multicolumn{3}{l|}{PPO\_jitter\_action} \\ \cmidrule(l){2-13} 
			& Win & Lost & Tie & Win & Lost & Tie & Win & Lost & Tie & Win & Lost & Tie \\ \midrule
			StaticAgent & 0.179 & 0.138 & 0.681 & 0.614 & 0.253 & 0.132 & 0.347 & 0.003 & 0.65 & 0.904 & 0.023 & 0.073 \\ \midrule
			SimpleAgent\_NoBomb & 0.373 & 0.366 & 0.260 & 0.425 & 0.446 & 0.128 & 0.583 & 0.081 & 0.335 & 0.778 & 0.111 & 0.111 \\ \midrule
			SimpleAgent & 0.347 & 0.379 & 0.273 & 0.426 & 0.45 & 0.123 & 0.622 & 0.079 & 0.298 & 0.778 & 0.088 & 0.135 \\ \midrule
			SimpleAgent\_NoBomb\_action & 0.230 & 0.586 & 0.183 & 0.271 & 0.641 & 0.086 & 0.507 & 016 & 0.333 & 0.681 & 0.19 & 0.129 \\ \midrule
			SimpleAgent\_action & 0.260 & 0.537 & 0.202 & 0.282 & 0.615 & 0.101 & 0.521 & 0.167 & 0.312 & 0.672 & 0.186 & 0.142 \\ \midrule
			PPO\_agent\_Cautious & 0.007 & 0.138 & 0.855 & 0.067 & 0.69 & 0.239 & 0.026 & 0.011 & 0.963 & 0.313 & 0.126 & 0.423 \\ \midrule
			Imitation\_Vanilla & 0.145 & 0.108 & 0.747 & 0.411 & 0.388 & 0.201 & 0.243 & 0.011 & 0.746 & 0.713 & 0.099 & 0.188 \\ \midrule
			Skynet & - & - & - & - & - & - & - & - & - & 0.451 & 0.126 & 0.423 \\ \bottomrule
		\end{tabular}
	\end{adjustbox}
	\caption{Results in 1000 games for PPO team (some games discarded due to fault after 12 steps).}
	\label{tab:results_ppo}
\end{table}
Fig. \ref{fig:effect_jit_act} explores the sensitivity of performance to the inclusion of jitter and action filters in each agent type. Specifically, we plot the change in wins, ties, losses (as a percentage of 1000 games) for the PPO agent against different opponents, when one or both filters are included. Jitter correction leads to fewer ties against all opponents, but increases both wins and losses. Action filter reduces losses against all opponents, but some of those losses are converted to ties. Using both jitter and action filters decreases losses as well as ties in all but one case, with more wins in all cases.
In Sec. \ref{sec:method}, we indicated that the curriculum of playing against gradually more difficult opponents leads to faster training than otherwise. In Fig.~\ref{fig:behaviour}, we provide some intuition behind this claim. The plots on the left are heatmaps of our agent's position while playing against different opponent types, and the plots on the right are heatmaps of bomb placement locations by our agent. All plots are aggregated over 50 games each, with our agent starting in the top left corner.

From Fig.~\ref{fig:exploration}, we observe that most extensive exploration happens when playing against StaticAgent and SimpleAgent\_NoBomb. This is because these two opponents are unable to leave their quadrants (cannot break wooden walls), which forces the PPO agent to hunt them down. On the other hand, the PPO agent can afford to be more conservative and wait for SimpleAgent or SimpleAgent\_action to engage it, requiring lower exploration. 

Fig.~\ref{fig:bomb_placement} shows similar behaviour, where bombs are placed in farther locations against StaticAgent and SimpleAgent\_NoBomb (bombs in its own quadrant are used to break wooden walls). The two more sophisticated opponents require more nuanced strategy, including (as seen from graphically rendered games) multiple bomb placement to create traps. However, were the PPO agent not trained against the simpler agents first, the exploration and bomb placement tendencies learnt through imitation would be forgotten very quickly (as seen in PPO\_Cautious).

\section{Conclusion}

We posit that the use of imitation followed by reinforcement learning is an effective way to reduce the training effort in Pommerman. Even if the expert policy for imitation is flawed, the agent is able to learn basic skills from it. Following this, reinforcement learning needs to be introduced gently (by training against simple opponents first) in order to retain the basic skills, while learning higher level skills against increasingly sophisticated opponents. 

\bibliography{ref.bib}
\bibliographystyle{IEEEtran}

\end{document}